\documentclass{pprai}

\usepackage[utf8]{inputenc}
\usepackage[T1]{fontenc}
\usepackage{graphicx}
\usepackage{amsthm}
\usepackage{txfonts}
\usepackage{url}
\usepackage{algpseudocode}

\title{A Python toolkit for dealing with Petri nets over ontological graphs}
\headtitle{A Python toolkit for dealing with Petri nets over ontological graphs}

\author{Krzysztof Pancerz$^{[0000-0002-5452-6310]}$}
\headauthor{K. Pancerz}
\affiliation{%
The John Paul II Catholic University of Lublin\\
Institute of Philosophy\\
Al. Rac{\l}awickie 14, 20-950 Lublin, Poland\\
kpancerz@kul.pl}

\keywords{modelling, simulation, Petri nets, OWL ontology, Python toolkit}

\begin{document}
\maketitle

\begin{abstract}
We present theoretical rudiments of Petri nets over ontological graphs as well as the designed and implemented Python toolkit for dealing with such nets. In Petri nets over ontological graphs, the domain knowledge is enclosed in a form of ontologies. In this way, some valuable knowledge (especially in terms of semantic relations) can be added to model reasoning and control processes by means of Petri nets. In the implemented approach, ontological graphs are obtained from ontologies built in accordance with the OWL 2 Web
Ontology Language. The implemented tool enables the users to define the structure and dynamics of Petri nets over ontological graphs.
\end{abstract}

\section{Introduction}

The main research goal is to use a new model of Petri nets covering semantic knowledge, called Petri nets over ontological graphs, to model reasoning and control processes. On the one hand, Petri nets are a powerful graphical and formal tool used to describe structures and dynamics of real-life systems. This tool was proposed by Carl Adam Petri in the early 1960s \cite{Petri1962}. On the other hand, ontologies specify the concepts and the relationships among them appearing in real-life areas \cite{Neches1991}. Therefore, in case of Petri nets over ontological graphs, the theoretical and graphic power of Petri nets is combined with the semantic power of ontologies. Currently, in the area of Petri nets, special attention in research and applications is focused on the so-called high level Petri nets \cite{Jensen1991} which enable us to obtain much more succinct and expressive descriptions than can be obtained by means of low level Petri nets (e.g., place-transition nets \cite{Reisig1985}). In the basic model of Petri nets, each place (corresponding to a state of a system) contains a dynamically varying number of small black dots, which are called tokens. An arbitrary distribution of tokens on the places is called a marking. Tokens can be interpreted as conditions, objects, items, etc. The Petri net dynamics is given by firing enabled transitions causing the movement of tokens through the net. There are many different classes of Petri nets extending the basic definition. Both low-level and high-level Petri nets are considered. In low-level nets, there is only one kind of tokens. In high-level nets, each token can carry complex information (for example, there are coloured tokens, fuzzy tokens, etc.). In the proposed approach, we intend to consider tokens as entities placed in semantic spaces (represented by ontologies, especially, OWL ontologies). Tokens are concepts which describe objects or phenomena. The conception of a new model of high-level Petri nets, i.e., Petri nets over ontological graphs, was presented in \cite{Szkola2017_IJCRS_2017}. The information carried by the token is much closer to the human perception. Therefore, analysis of such models is easier. Moreover, it enables us to define the conditions for firing transitions in a coherent way on the basis of linguistic semantics of tokens. The OWL ontologies lead us to conception of two types of models of Petri nets over ontological graphs. This conception was presented in \cite{Pancerz_CSP_2017}. The first model is a conceptually marked Petri net over ontological graphs. In this model, tokens are concepts from ontological graphs associated with places of a Petri net. Dynamics of such Petri nets determines a flow of concepts. The second model is an instancely marked Petri net over ontological graphs. In this model, tokens are instances of concepts from ontological graphs associated with places of a Petri net. Dynamics of such Petri nets determines a flow of instances. 

\section{Rudiments}

Formally, the ontology can be represented by means of graph structures (cf. \cite{Pancerz2012a}). In this case, the graph representing the ontology $\mathcal{O}$ is called the ontological graph. It is a tuple including: $\mathcal{C}$ - the finite set of nodes representing concepts in the ontology $\mathcal{O}$, $E$ - the finite set of edges representing semantic relations between concepts, $\mathcal{R}$ - the family of semantic descriptions (in a natural language) of types of relations (represented by edges) between concepts, $\rho$ - the function assigning a semantic description of the relation to each edge. Ontological graphs can be obtained from ontologies built in accordance with the OWL 2 Web Ontology Language (shortly OWL 2). An OWL ontology consists of three components: classes, individuals, and properties. Classes are representations of concepts, individuals are instances of classes, properties are binary relations on individuals. For ontological graphs obtained from OWL ontologies, $INST(\mathcal{C})$ is a set of all instances of the concepts from the set $\mathcal{C}$. 

In general, ontologies model varied semantic relations between concepts. Let $c$ and $c'$ be concepts and let $i$ be an instance in a given ontology. Our attention is focused on three basic semantic relations, namely: \textbf{EQUIV-TO} (if $c$ EQUIV-TO $c'$, then $c$ is a synonym of $c'$), \textbf{SUBCLASS-OF}, also known as IS-A (if $c$ SUBCLASS-OF $c'$, then $c$ is a kind of $c'$), \textbf{INSTANCE-OF} (if $i$ INSTANCE-OF $c$, then $i$ is an instance of $c$). According to description logics (cf. \cite{Baader2004}), we consider two distinguished concepts with useful applications, namely the top concept $\top$ (i.e., a concept with every individual as an instance) and the bottom concept $\bot$ (i.e., an empty concept with no individuals as instances). Further, $\epsilon$ will be an instance of the bottom concept $\bot$. We can build some formulas over the set of concepts from the ontological graph $OG$. Formulas are written according to the syntax used in the designed and implemented Python toolkit. In the presented approach, we will use formulas as follows: $c$ that means $c \in \mathcal{C}$, $\{c\}$ that means $\{c\}$, $[c]$ that means $\{c' \in \mathcal{C}: c'=c \mbox{ or } c' \mathop{EQUIV-TO} c\}$, as well as $<c>$ that means $\{c' \in \mathcal{C}: c' \neq \bot \mbox{ and }  c' \mathop{SUBCLASS-OF} c\}$.

\begin{table}[!bht]
\begin{center}
\begin{footnotesize}
\begin{tabular}{|c|c|c|}
\hline
Item & Description & Remarks \\
\hline \hline
$Pl$ & The finite set of places & \\
\hline
$Tr$ & The finite set of transitions &  \\
\hline
$\{OG\}_{p \in Pl}$ & The family of ontological &  $OG_p=(\mathcal{C}_p, E_p, \mathcal{R}_p, \rho_p)$  \\
 & graphs associated with places & for each $p \in Pl$ \\
\hline
$Arc_{in}$ & The set of input arcs & $Arc_{in} \subseteq Pl \times Tr$ \\
\hline
$Arc_{out}$ & The set of output arcs & $Arc_{out} \subseteq Tr \times Pl$ \\
\hline
\multicolumn{3}{|c|}{For $CMPNOG$:}  \\
\hline
$Form_{in}$ & The input arc formula function & $||Form_{in}(p,t)||_{OG_p} \subseteq \mathcal{C}_p$ \\
 & & for each $(p,t) \in Arc_{in}$ \\
\hline
$Form_{out}$ & The output arc formula function & $||Form_{out}(t,p)||_{OG_p} \in \mathcal{C}_p$ \\
 & & for each $(t,p) \in Arc_{out}$ \\
 \hline
$Mark_0$ & The initial marking function & $Mark_0(p) \in \{\bot\} \cup \mathcal{C}_p$ \\
 & & for each $p \in Pl$ \\
\hline
\multicolumn{3}{|c|}{For $IMPNOG$:}  \\
\hline
$Form_{in}$ & The input arc formula function & $||Form_{in}(p,t)||_{OG_p} \in \mathcal{C}_p$ \\
 & & for each $(p,t) \in Arc_{in}$ \\
\hline
$Form_{out}$ & The output arc formula function & $||Form_{out}(t,p)||_{OG_p} \in INST(\mathcal{C}_p)$ \\
 & & for each $(t,p) \in Arc_{out}$ \\
 \hline
$Mark_0$ & The initial marking function & $Mark_0(p) \in \{\epsilon\} \cup  INST(\mathcal{C}_p)$ \\
 & & for each $p \in Pl$ \\
\hline
\end{tabular}
\end{footnotesize}
\end{center}
\label{Tab:PNOGs}
\caption{Items of Petri nets over ontological graphs.}
\end{table}

\begin{table}[!bht]
\begin{center}
\begin{footnotesize}
\begin{tabular}{|c|c|}
\hline
Petri net & Conditions \\
\hline \hline
$CMPNOG$ & $Mark(p) \in ||Form_{in}(p,t)||_{OG_p}$  for all $p \in Pl$ such that $(p,t) \in Arc_{in}$ \\
 & $Mark(p)=\bot$ for all $p \in Pl$ such that $(t,p) \in Arc_{out}$ \\
\hline
$IMPNOG$ & $Mark(p) \in ||Form_{in}(p,t)||_{OG_p}$ for all $p \in Pl$ such that $(p,t) \in Arc_{in}$ \\
  & $Mark(p)=\epsilon$ for all $p \in Pl$ such that $(t,p) \in Arc_{out}$ \\
\hline
\end{tabular}
\end{footnotesize}
\end{center}
\label{Tab:enabling}
\caption{Conditions for transitions to be enabled.}
\end{table}

\begin{table}[!bht]
\begin{center}
\begin{footnotesize}
\begin{tabular}{|c|c|c|}
\hline
Petri net & New marking \\
\hline \hline
$CMPNOG$ & $Mark'(p)=\bot$ if $p \in Pl$ and $(p,t) \in Arc_{in}$ \\
 & $Mark'(p)=||Form_{out}(t,p)||_{OG_p}$ if $p \in Pl$ and $(t,p) \in Arc_{out}$ \\
 & $Mark'(p)=Mark(p)$, otherwise \\
\hline
$IMPNOG$ & $Mark'(p)=\epsilon$ if $p \in Pl$ and $(p,t) \in Arc_{in}$ \\
 & $Mark'(p)=||Form_{out}(t,p)||_{OG_p}$ if $p \in Pl$ and $(t,p) \in Arc_{out}$ \\
 & $Mark'(p)=Mark(p)$, otherwise \\
\hline
\end{tabular}
\end{footnotesize}
\end{center}
\label{Tab:firing}
\caption{A new marking after firing a transition.}
\end{table}

A conceptually marked Petri net $CMPNOG$ over ontological graphs and an instancely marked Petri net $IMPNOG$ over ontological graphs are tuples with items listed in Table \ref{Tab:PNOGs} (cf. \cite{Szkola2017_IJCRS_2017}). A transition $t \in Tr$ is said to be enabled if and only if proper conditions given in Table \ref{Tab:enabling} are satisfied. Firing an enabled transition $t$ causes the movement of tokens as it is shown in Table \ref{Tab:firing}.

\section{Example}

\begin{figure}[ht] 
\centering
\includegraphics[width=1.0\textwidth]{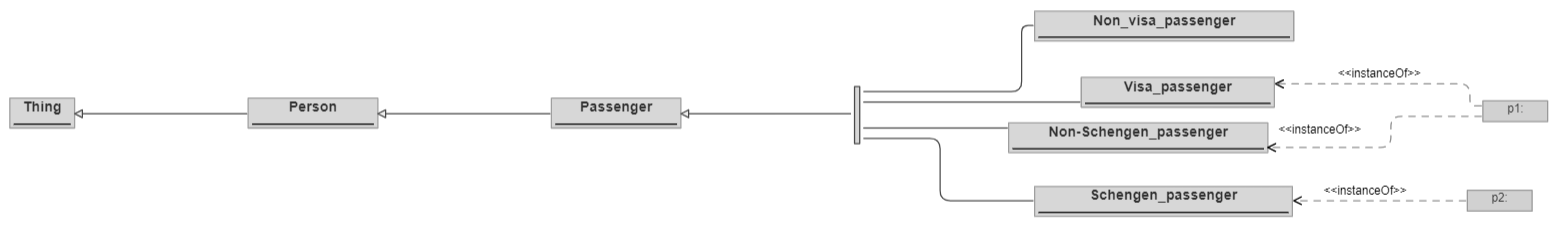}
\caption{The ontological graph $og_1$ assigned to place $pl_1$.}
\label{Fig:OG_passengers}
\end{figure}

\begin{figure}[ht] 
\centering
\includegraphics[width=1.0\textwidth]{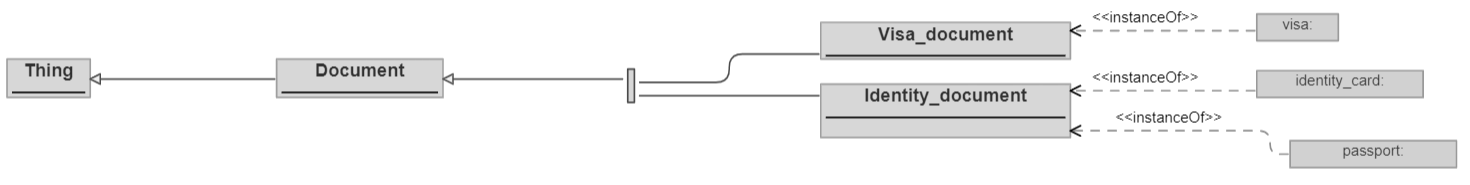}
\caption{The ontological graph $og_2$ assigned to places $pl_2$ and $pl_3$.}
\label{Fig:OG_documents}
\end{figure}

The main idea of instancely marked Petri nets over ontological graphs is shown in a simple example as follows. The ontological graphs $og_1$ and $og_2$ shown in Figures \ref{Fig:OG_passengers} and \ref{Fig:OG_documents} are assigned to places $pl_1$ as well as $pl_2$ and $pl_3$, respectively. 

\begin{figure}[ht] 
\centering
\includegraphics[width=0.8\textwidth]{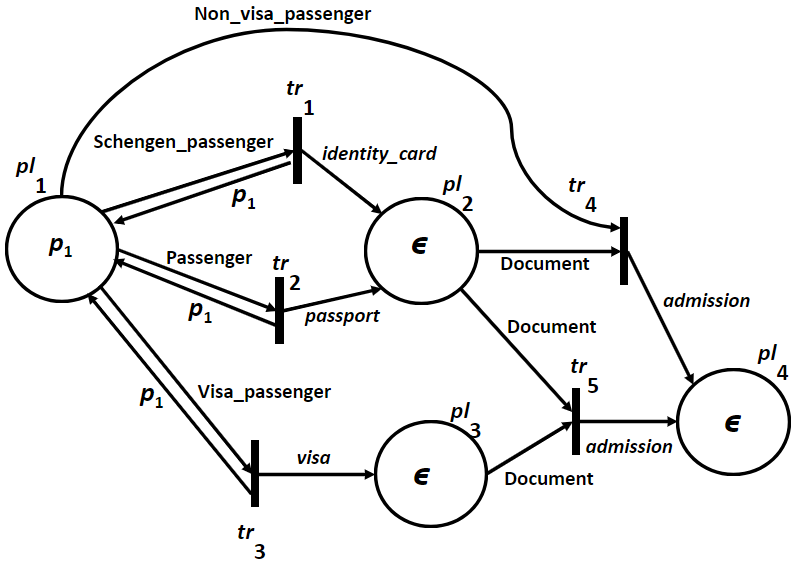}
\caption{A simple example of an instancely marked Petri net (the initial marking).}
\label{Fig:PNOG_1}
\end{figure}  

\begin{figure}[ht] 
\centering
\includegraphics[width=0.8\textwidth]{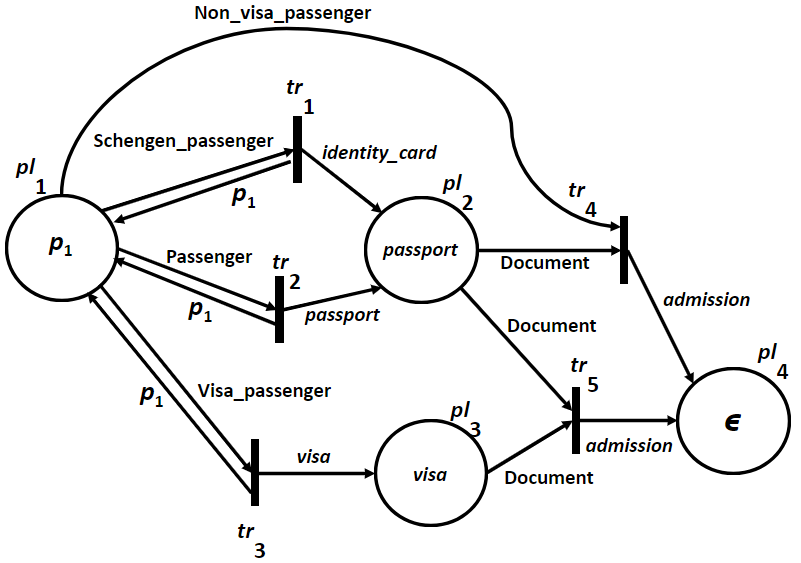}
\caption{A simple example of an instancely marked Petri net (the marking after firing transitions $tr_2$ and $tr_3$).}
\label{Fig:PNOG_2}
\end{figure} 

\begin{figure}[ht] 
\centering
\includegraphics[width=0.8\textwidth]{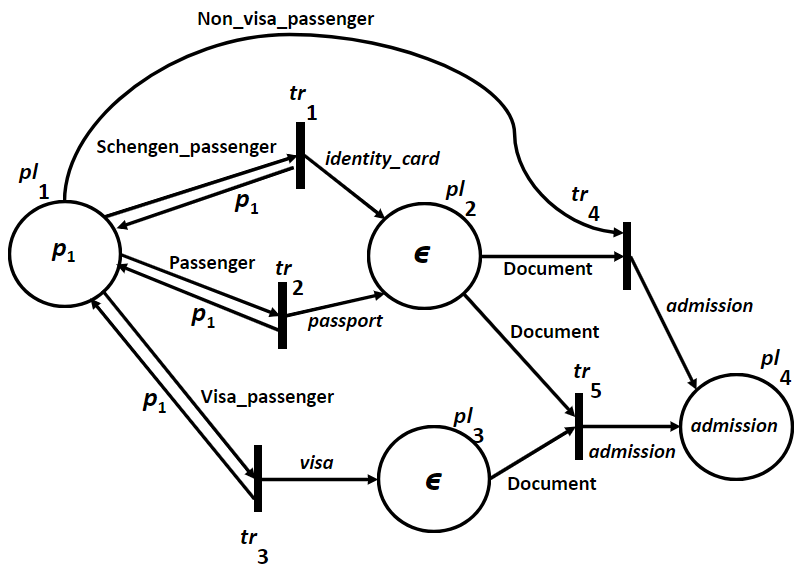}
\caption{A simple example of an instancely marked Petri net (the marking after firing transition $tr_5$).}
\label{Fig:PNOG_3}
\end{figure} 

Input arcs are described by formulas which are classes from the ontological graphs $og_1$ and $og_2$. If $p_1$ INSTANCE-OF $Visa\_passenger$ (see the initial marking shown in Figure \ref{Fig:PNOG_1}), then both transitions $tr_2$ and $tr_3$ are enabled to fire. It is worth noting that $p_1$ INSTANCE-OF $Passenger$ holds because $Visa\_passenger$ SUBCLASS-OF $Passenger$. After firing $tr_2$ and $tr_3$, we obtain a new marking shown in Figure \ref{Fig:PNOG_2}. The instance $passport$ is sent (as a token) to $pl_2$ and the instance $visa$ is sent (as a token) to $pl_3$. At this state of the Petri net, transition $tr_5$ is enabled to fire. The final marking (after firing $tr_5$) is shown in Figure \ref{Fig:PNOG_3}.

The Python code that uses objects from the implemented Python tool is as follows. In the toolkit, we have used the \textit{owlready2} package to process OWL structures (describing ontological graphs). We assume that arcs (both input and output) with assigned formulas are given in the form of matrices with rows labelled with places and columns labelled with transitions. $EPS$ is a constant representing the instance of the bottom concept $\bot$.  

\begin{scriptsize}
\begin{verbatim}
import owlready2
ont_world_1 = owlready2.World()
og1=ont_world_1.get_ontology('OG_passengers.owl').load()
ont_world_2 = owlready2.World()
og2=ont_world_2.get_ontology('OG_documents.owl').load()
ontological_graphs=[og1,og2]
pl1=PNOG_place('pl1')
pl2=PNOG_place('pl2')
pl3=PNOG_place('pl3')
pl4=PNOG_place('pl4')
tr1=PNOG_transition('tr1')
tr2=PNOG_transition('tr2')
tr3=PNOG_transition('tr3')
tr4=PNOG_transition('tr4')
tr5=PNOG_transition('tr5')
places=[pl1,pl2,pl3,pl4]
transitions=[tr1,tr2, tr3, tr4, tr5]
input_arcs=  
   [['Schengen_passenger','Passenger','Visa_passenger','Non_visa_passenger',None],
   [None,None,None,'Document','Document'],
   [None,None,None,None,'Document],
   [None,None,None,None,None]]
output_arcs=
   [['p1','p1','p1',None,None],
   ['identity_card','passport',None,None,None],
   [None,None,'visa',None,None],
   [None,None,None,'admission','addmision']]
m0=['p1', EPS, EPS, EPS]
net=PNOG(places, transitions, ontological_graphs, input_arcs, output_arcs, m0)
\end{verbatim}
\end{scriptsize}

\section{Conclusions}

Petri nets over ontological graphs can be used wherever linguistic concepts are used, taking into account, among others, the hierarchy of their meanings. In general, they can be used to describe structures and behaviours of business processes, reasoning processes, control processes, etc. The outline of the work plan for the future is as follows: designing and implementing procedures for finding occurrence graphs and place and transitions invariants, as well as defining a dedicated programming language to specify declarations and net inscriptions.




\bibliography{PP_RAI_2024_KP}

\begin{thebibliography}{1}
\providecommand{\url}[1]{\texttt{#1}}
\providecommand{\urlprefix}{URL }
\expandafter\ifx\csname urlstyle\endcsname\relax
  \providecommand{\doi}[1]{doi:\discretionary{}{}{}#1}\else
  \providecommand{\doi}{doi:\discretionary{}{}{}\begingroup
  \urlstyle{rm}\Url}\fi

\bibitem{Petri1962}
Petri, C.
\newblock Kommunikation mit automaten.
\newblock Schriften des {IIM} nr. 2, Institut f{\"u}r Instrumentelle
  Mathematik, Bonn, 1962.

\bibitem{Neches1991}
Neches, R., Fikes, R., Finin, T., Gruber, T., Patil, R., Senator, T., and
  Swartout, W.
\newblock Enabling technology for knowledge sharing.
\newblock \emph{AI Magazine}, 12(3):36--56, 1991.

\bibitem{Jensen1991}
Jensen, K. and Rozenberg, G., editors.
\newblock \emph{High-level {Petri} Nets: Theory and Application}.
\newblock Springer-Verlag, Berlin Heidelberg, 1991.

\bibitem{Reisig1985}
Reisig, W.
\newblock \emph{Petri Nets: An Introduction}.
\newblock Springer, Berlin, 1985.

\bibitem{Szkola2017_IJCRS_2017}
Szko{\l}a, J. and Pancerz, K.
\newblock Petri nets over ontological graphs: Conception and application for
  modelling tasks of robots.
\newblock In L.~Polkowski et~al., editors, \emph{Rough Sets}, volume 10313 of
  \emph{LNAI}, pages 207--214. Springer International Publishing, Cham, 2017.

\bibitem{Pancerz_CSP_2017}
Pancerz, K., Grochowalski, P., and Paja, W.
\newblock Two simple models of petri nets over ontological graphs.
\newblock In P.~van Beek, editor, \emph{Proc. of CS\&P'2017}. 2017.

\bibitem{Pancerz2012a}
Pancerz, K.
\newblock Toward information systems over ontological graphs.
\newblock In J.~Yao et~al., editors, \emph{Rough Sets and Current Trends in
  Computing}, volume 7413 of \emph{LNAI}, pages 243--248. Springer-Verlag,
  Berlin Heidelberg, 2012.

\bibitem{Baader2004}
Baader, F., Horrocks, I., and Sattler, U.
\newblock Description logics.
\newblock In S.~Staab and R.~Studer, editors, \emph{Handbook on Ontologies},
  pages 3--28. Springer, Berlin Heidelberg, 2004.

\end{thebibliography}
\bibliographystyle{pprai}

\end{document}